\pdfoutput=1

\documentclass[11pt]{article}

\usepackage[]{ACL2023}

\usepackage{times}
\usepackage{latexsym}

\usepackage[T1]{fontenc}

\usepackage[utf8]{inputenc}

\usepackage{microtype}

\usepackage{inconsolata}

\usepackage{enumitem}
\setenumerate[1]{itemsep=0pt,partopsep=0pt,parsep=\parskip,topsep=5pt}
\setitemize[1]{itemsep=0pt,partopsep=0pt,parsep=\parskip,topsep=2pt}
\setdescription{itemsep=0pt,partopsep=0pt,parsep=\parskip,topsep=5pt}

\usepackage{graphicx}
\usepackage{algorithm}
\usepackage{algorithmic}

\usepackage{multirow}

\usepackage{amsmath,amsfonts,bm}

\def\eqref#1{equation~\ref{#1}}

\def\1{\bm{1}}

\def\ra{{\textnormal{a}}}

\def\ermK{{\textnormal{K}}}

\def\ermN{{\textnormal{N}}}

\def\ermT{{\textnormal{T}}}

\def\vw{{\bm{w}}}

\def\mW{{\bm{W}}}

\DeclareMathAlphabet{\mathsfit}{\encodingdefault}{\sfdefault}{m}{sl}
\SetMathAlphabet{\mathsfit}{bold}{\encodingdefault}{\sfdefault}{bx}{n}

\def\gC{{\mathcal{C}}}
\def\gD{{\mathcal{D}}}
\def\gE{{\mathcal{E}}}

\def\gG{{\mathcal{G}}}

\def\gL{{\mathcal{L}}}

\def\gP{{\mathcal{P}}}

\def\gR{{\mathcal{R}}}

\def\gX{{\mathcal{X}}}
\def\gY{{\mathcal{Y}}}

\def\sA{{\mathbb{A}}}

\def\sD{{\mathbb{D}}}

\def\sG{{\mathbb{G}}}

\def\sK{{\mathbb{K}}}

\def\sN{{\mathbb{N}}}

\def\sP{{\mathbb{P}}}

\def\sT{{\mathbb{T}}}

\def\sY{{\mathbb{Y}}}

\usepackage{mathtools}
\DeclarePairedDelimiter{\abs}{\lvert}{\rvert}
\usepackage{xcolor}

\title{Boosting Event Extraction with Denoised Structure-to-Text Augmentation}

\author{Bo Wang\textsuperscript{1,2,3} , Heyan Huang\textsuperscript{1,2,3}\thanks{\ \ Corresponding author.} , Xiaochi Wei \textsuperscript{5} , Ge Shi\textsuperscript{4},
     {\bf Xiao Liu\textsuperscript{1,2,3}},\\  {\bf Chong Feng \textsuperscript{1,2,3} } , {\bf Tong Zhou \textsuperscript{4}} , {\bf Shuaiqiang Wang\textsuperscript{5}} , {\bf Dawei Yin\textsuperscript{5}}\\       
        \textsuperscript{1}School of Computer Science and Technology, Beijing Institute of Technology 
        \\ \textsuperscript{2}Key Lab of IIP\&IS, Ministry of Industry and Information Technology, China
        \\ \textsuperscript{3}Southeast Academy of Information Technology, Beijing Institute of Technology 
        \\ \textsuperscript{4}Faculty of Information Technology, Beijing University of Technology \ \  
         \textsuperscript{5}Baidu Inc.
        \\ \texttt{\{bwang,hhy63\}@bit.edu.cn}
        }

\begin{document}
\maketitle
\begin{abstract}
Event extraction aims to recognize pre-defined event triggers and arguments from texts, which suffer from the lack of high-quality annotations.
In most NLP applications, involving a large scale of synthetic training data is a practical and effective approach to alleviate the problem of data scarcity.
However, when applying to the task of event extraction, recent data augmentation methods often neglect the problem of grammatical incorrectness, structure misalignment, and semantic drifting, leading to unsatisfactory performances.
In order to solve these problems, we propose a denoised structure-to-text augmentation framework for event extraction (\textsc{DAEE}), which generates additional training data through the knowledge-based structure-to-text generation model and selects the effective subset from the generated data iteratively with a deep reinforcement learning agent.
Experimental results on several datasets demonstrate that the proposed method generates more diverse text representations for event extraction and achieves comparable results with the state-of-the-art.
\end{abstract}
\section{Introduction}

\begin{figure}[t]
\centering
\setlength{\belowcaptionskip}{-0.6cm}
\includegraphics[width=0.95\columnwidth]{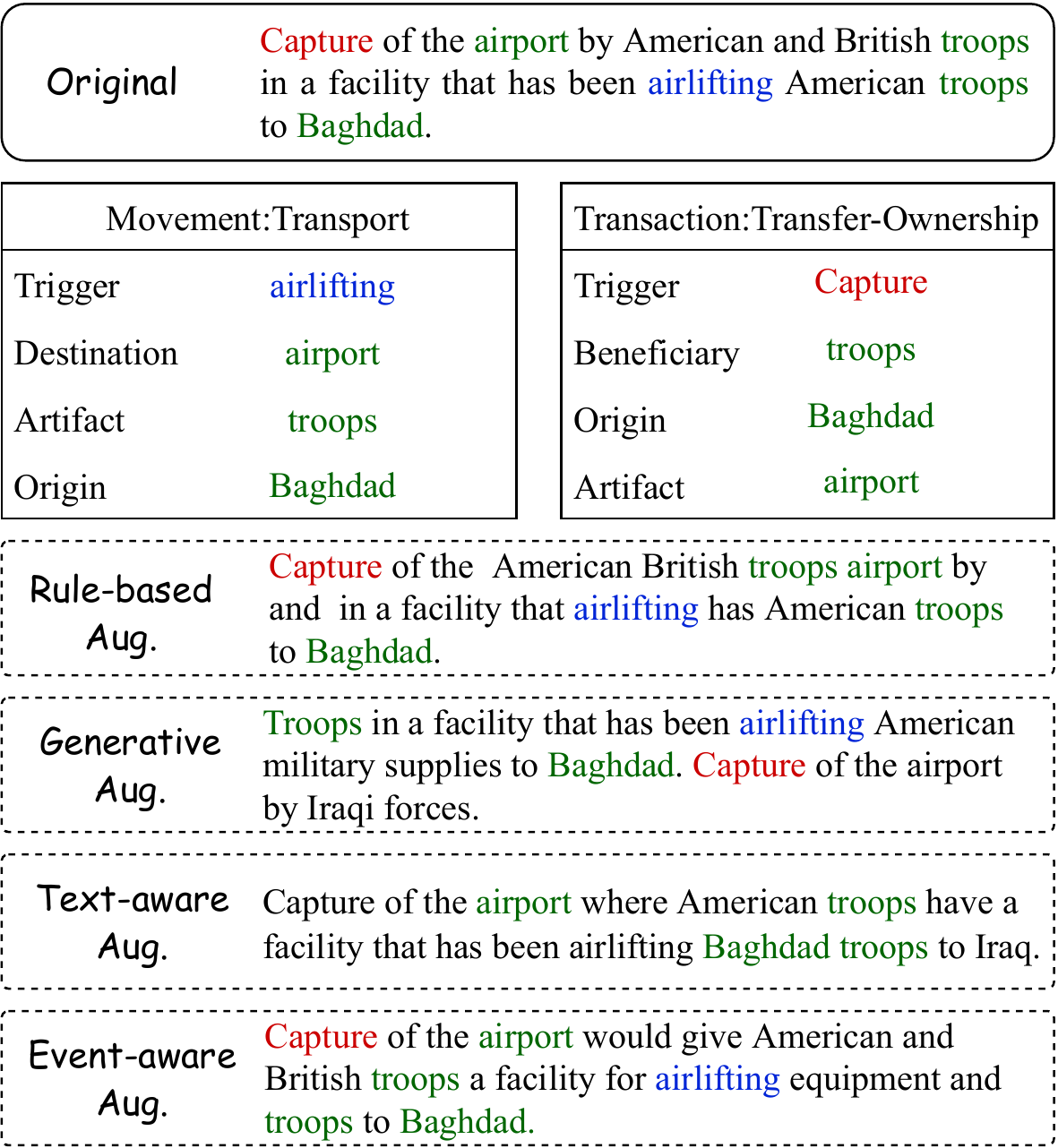} 
\caption{Example of text data augmentation methods.}
\label{fig: ex}
\end{figure}

Event extraction is an essential yet challenging task for natural language understanding.
Given a piece of text, event extraction systems discover the event mentions and then recognize event triggers and their event arguments according to pre-defined event schema ~\cite{DBLP:conf/lrec/DoddingtonMPRSW04,ahn-2006-stages}. 
As shown in Figure~\ref{fig: ex}, the sentence ``\textit{Capture of the airport by American and British troops in a facility that has been airlifting American troops to Baghdad.}" contains two events, 
a \texttt{Movement:Transport} event triggered by ``\textit{airlifting}” and a \texttt{Transaction:Transfer-Ownership} event triggered by ``\textit{Capture}”. 
In the \texttt{Movement:Transport} event, three event roles are involved, i.e., \texttt{Artifact}, \texttt{Destination}, and \texttt{Origin}, and their arguments are \textit{troops}, \textit{airports}, and \textit{Baghdad}, respectively. 
As to the \texttt{Transaction:Transfer-Ownership} event, the event roles are \texttt{Beneficiary}, \texttt{Origin}, and \texttt{Artifact}. Accordingly, the arguments are \textit{troops}, \textit{Baghdad}, and \textit{airports}.

Traditional event extraction methods regard the task as a trigger classification sub-task and several arguments classification sub-tasks~\cite{du-cardie-2020-event,liu-etal-2020-event,lin-etal-2020-joint,zhang-ji-2021-abstract,nguyen-etal-2021-cross,nguyen-etal-2022-joint,nguyen-etal-2022-learning}, while some of the recent research casting the task as a sequence generation problem~\cite{DBLP:conf/iclr/PaoliniAKMAASXS21,li-etal-2021-document,hsu-etal-2022-degree,xiaoliu2023}.
Compared with classification-based methods, the latter line is more data-efficient and flexible.
Whereas, the data containing event records are scarce, and the performance is influenced by the amount of data as the results shown in~\citet{hsu-etal-2022-degree}.

As constructing large-scale labeled data is of great challenge, data augmentation plays an important role here to alleviate the data deficient problem.
There are three main augmentation methods, i.e., Rule-based augmentation method~\cite{DBLP:conf/emnlp/WeiZ19,DBLP:conf/coling/DaiA20}, generative method~\cite{DBLP:conf/iccS/WuLZHH19,DBLP:journals/tacl/KumarAVT20,DBLP:conf/aaai/Anaby-TavorCGKK20,wei-zou-2019-eda,DBLP:conf/emnlp/NgCG20}, and text-aware method~\cite{,ding-etal-2020-daga}. However, they have different drawbacks.
1) \textbf{Grammatical Incorrectness.} Rule-based methods expand the original training data using automatic heuristic rules, such as randomly synonyms replacement, which effectively creates new training instances. As the example of \textit{Rule-based Aug} illustrated in Figure~\ref{fig: ex}, these processes may distort the text, making the generated syntactic data grammatically incorrect.
2) \textbf{Structure Misalignment.} Triggers and arguments are key components of event records, whether for both the original one and the augmented one. Nonetheless, triggers and arguments may not always exist in previous augmentation methods. As the example of \textit{Generative Aug} illustrated in Figure~\ref{fig: ex}, even though the meaning of the generated augmented sentence is quite similar to the original one, the important argument ``\textit{airport}" is missing. This may mislead the model to weaken the recognition of the \textsc{Destination} role.
3) \textbf{Semantic Drifting.} 
Another important aspect of data augmentation is semantic alignment. The generated text needs to express the original event content without semantic drifting. However, this problem is commonly met in the \textit{Text-aware Aug} method. As the example illustrated in Figure~\ref{fig: ex}, the sentence completely contains all the triggers and arguments. But instead of \textit{Baghdad}, \textit{Iraq} is regarded as the \textsc{Origin} in generated sentences, which may confuse the model to recognize the correct \textsc{Origin} role.

In order to solve the aforementioned problem when applying data augmentation to event extraction, we proposed a denoised structure-to-text augmentation framework for event extraction (\textsc{DAEE}).
For structure misalignment problems, a knowledge-based structure-to-text generation model is proposed. It is equipped with an additional argument-aware loss to generate augmentation samples that exhibit features of the target event.
For the \textbf{Semantic Drift} problem, we designed a deep reinforcement learning~(RL) agent. It distinguishes whether the generated text expresses the corresponding event based on the performance variation of the event extraction model. 
At the same time, the agent further guides the generative model to pay more attention to the samples with the \textbf{Structure Misalignment} and \textbf{Grammatical Incorrectness} problems and thus affords the \textit{Event-aware Aug} text that both contain important elements and represent appropriate semantics.
Intuitively, our agent is able to select effective samples from the combination of generated text and its event information to maximize the reward based on the event extraction model. 

The key contributions of this paper are threefold:
\begin{itemize}
\item[$\bullet$] We proposed a denoised structure-to-text augmentation framework. It utilizes an RL agent to select the most effective subset from the augmented data to enhance the quality of the generated data.
\item[$\bullet$] Under the proposed framework, a knowledge-based structure-to-text generation model is proposed to satisfy the event extraction task, which generates high-quality training data containing corresponding triggers and arguments.
\item[$\bullet$]Experimental results on widely used benchmark datasets prove that the proposed method achieves superior performance over state-of-the-art event extraction methods on one dataset and comparable results on the other datasets.
\end{itemize}
\section{Related Work}
\subsection{Event Extraction}
Many existing methods use classification-based models to extract events~\cite{nguyen-etal-2016-joint-event, wang-etal-2019-hmeae, yang-etal-2019-exploring, wadden-etal-2019-entity,liu-etal-2018-jointly}.
And some global features are introduced to make an enhancement for joint inference~\cite{lin-etal-2020-joint, li-etal-2013-joint, yang-mitchell-2016-joint}.
With the large-scale use of PLMs, some of the researchers dedicated to developing generative capabilities for PLMs in event extraction, i.e., transforming into translation tasks~\cite{DBLP:conf/iclr/PaoliniAKMAASXS21}, generating with constrained decoding methods~\cite{lu-etal-2021-text2event}, and template-based conditional generation~\cite{li-etal-2021-document,hsu-etal-2022-degree,DBLP:conf/acl/LiuHSW22,DBLP:conf/acl/DuLJ22}.
Compare with the above method directly uses a limited number of the training set, we use a denoised structure-to-text augmentation method to alleviate the problem of insufficient data.

\subsection{Data Augmentation}
Rather than starting from an existing example and modifying it, some model-based data augmentation approaches directly estimate a generative process produce new synthetic data by masking randomly chosen words from the training set and sample from it~\cite{DBLP:conf/aaai/Anaby-TavorCGKK20,DBLP:conf/coling/HouLCL18,DBLP:conf/acl/XiaKAN19,DBLP:conf/iccS/WuLZHH19,DBLP:journals/tacl/KumarAVT20}.
Other research design prompt~\cite{DBLP:conf/acl/0003XSHTGJ22,DBLP:journals/corr/abs-2109-09193} or use conditional generation~\cite{ding-etal-2020-daga} for the data augmentation.
However, the above methods are mainly applied to generation tasks or comprehension tasks with simpler goals, such as text classification. 
When faced with complex structured extraction tasks, post-processing screening becomes a cumbersome problem. 
Inspired by RL, we use a policy model to automatically sift through the generated data for valid and semantically consistent samples.

\section{Method}
\label{section:A}
In this paper, we focus on generating the additional training set from structured event records for augmentation.
Previous augmentation methods usually have \textbf{Structure Misalignment} and \textbf{Grammatical Incorrectness}, and \textbf{Semantic Drifting} problems as mentioned in the introduction.
Instead, we introduce a policy-based RL strategy to select intact augmentation sentences.

\subsection{Task Definition}
In the generation-based event extraction task, the extraction process is divided into several subtasks according to event types $\gE$. 
For each event type $e \in \gE$, the purpose of the event extraction model is to generate $\gY_e$ according to the predefined prompt $\gP_e$ and context $\gC$, where $\gY_e$ is the answered prompts containing extracted event records.
Except for the original data $\sT_o$, we use a policy model as RL agent to select the effective subset $\sP_i$ from the generated data $\sG_i$ in the $i$-th epoch, thus improving the data efficiency by filtering the generated samples.

\subsection{Framework}
\begin{figure}[t]
\centering
\setlength{\belowcaptionskip}{-0.6cm}
\includegraphics[width=0.48\textwidth]{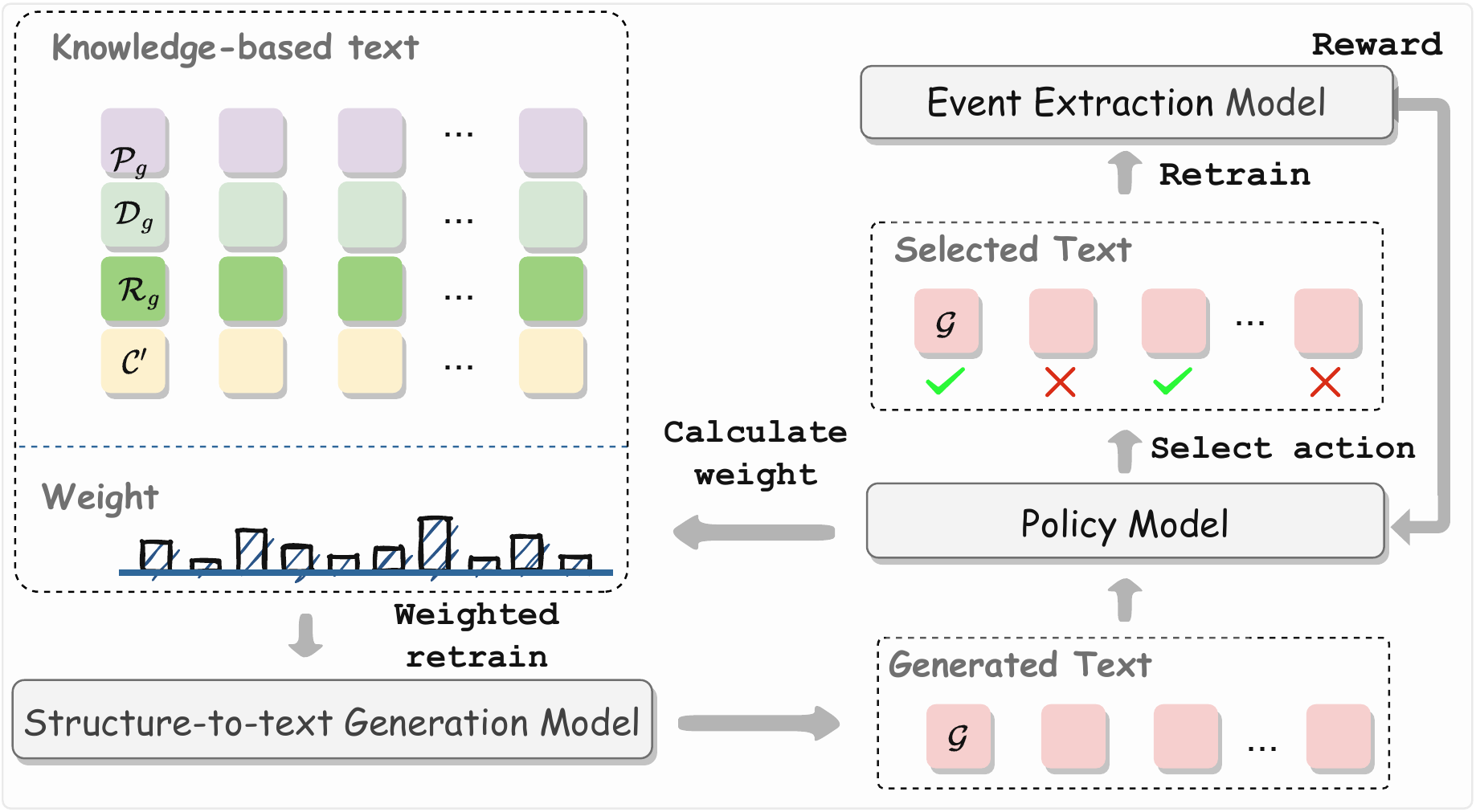}
\caption{The proposed policy-based RL framework.}
\label{fig: rl}
\end{figure}

Our proposed denoised structure-to-text augmentation framework is mainly composed of the event extraction model, structure-to-text generation model, and policy model.
As the policy-based RL process shown in Figure \ref{fig: rl}, the event record is first fed into the structure-to-text generation model to obtain the additional training data. Then they are filtered according to the action selected by the policy-based agent.
Thus, we obtain the denoised augmentation training data for event extraction model.
We use the filtered training data to retrain the event extraction model and the enhancement of the F1 score is regarded as a reward to retrain the policy model.
The guidance of the event extraction model further helps the policy model select efficient samples.
Finally, the generation model is retrained according to the weighted training data, and the weight is the removing action probability calculated by the retrained policy model.
The retraining captain the generation model produces superior-quality sentence and consequently help the other components.
The components of our proposed method will be described in the following.
\subsection{Reinforcement Learning components} 
The definitions of the fundamental components are introduced in the following.
The \textbf{States} include the information from the current sentence and the corresponding golden event records. These two parts are both converted to the sentence vector through PLMs for the decision of action. 
We update states after re-generate the text guided by the previous action probability.
At each iteration, the \textbf{Actions} decided by the policy model is whether to remove or retain the generated instance according to whether the sentences generated do express the corresponding event records.
We use the enhancement of the F1 score as the \textbf{Rewards} for the actions decided by the policy model. 
Specifically, the F1 score of argument classification $F_i$ at $i$-th epoch on the development set is adopted as the performance evaluation criterion.
Thus, the reward $\gR_i$ can be formulated as the difference between the adjacent epochs: 
\begin{align}
    \gR_i = \alpha(F_i - F_{i-1}),
\end{align}
where $\alpha$ is a scaling factor to convert the reward into a numeric result for RL agent.

\begin{figure}[t]
\centering
\includegraphics[width=0.44\textwidth]{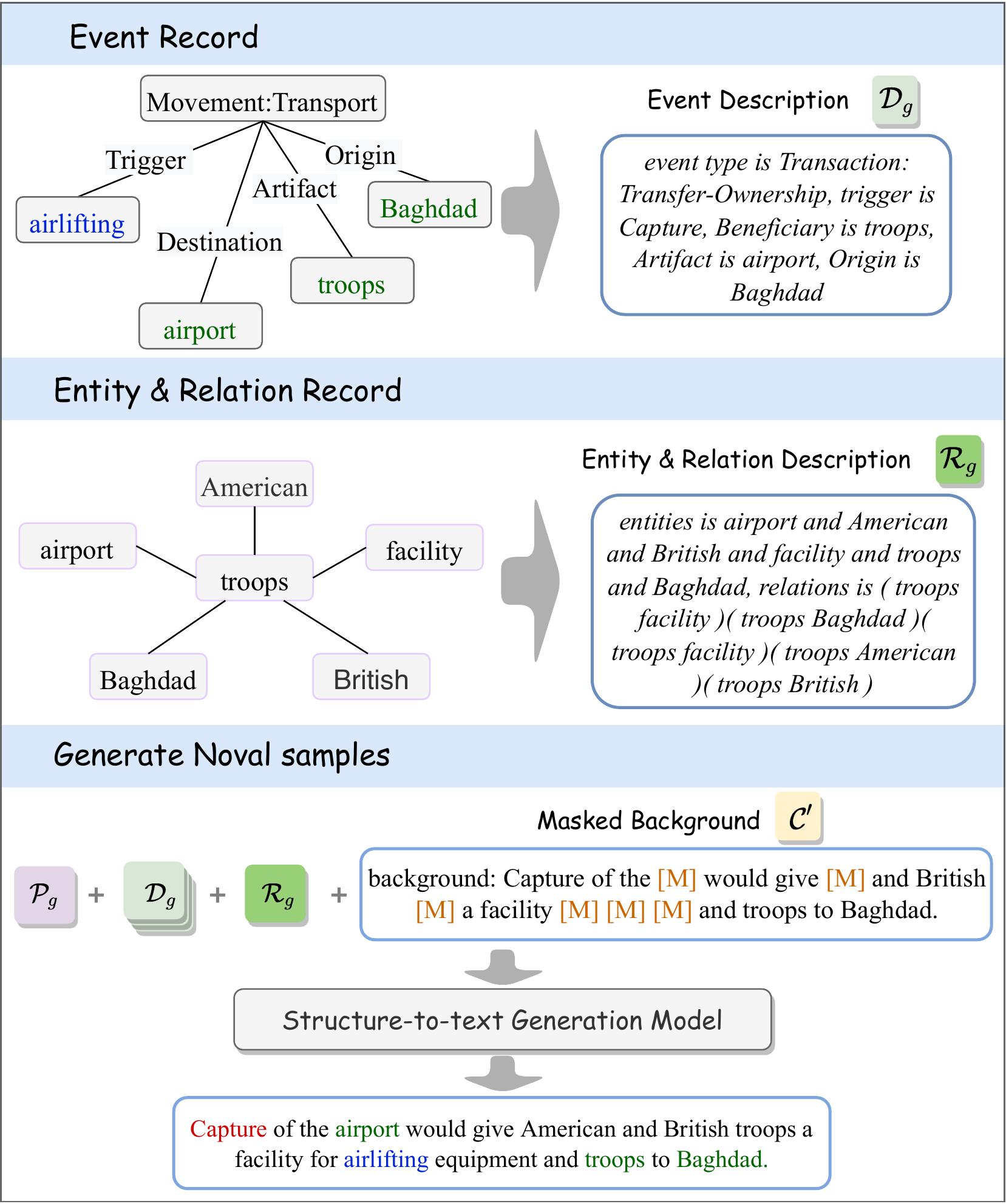} 
\caption{Example of structured information representations and structure-to-text generation.}
\label{fig: info2text}
\end{figure}
\subsubsection{Event Extraction Model}

We use the generation-based method \textsc{GTEE-Base}~\cite{DBLP:conf/acl/LiuHSW22} with the trained irrelevance classifiers as the event extraction model. 
The event extraction model is based on BART~\cite{lewis-etal-2020-bart}, the entire probability $p(\gY_e \mid \gX_e)$ is calculated through formulated input $\gX_e  =\left[\gP_{e} ; \texttt{[SEP]} ; \gC \right]$, where $[\ ;\ ]$ 
denotes the sequence concatenation operation, and \texttt{[SEP]} is the corresponding separate marker.
Following \cite{li-etal-2021-document} to reuse the predefined argument templates, the prompt $\gP_{e}$ contains the type instruction and the template, and the event records are parsed by template matching and slot mapping according to their own event description template.

\subsubsection{Structure-to-text Generation Model}
As to the structure-to-text generation model, T5~\cite{DBLP:journals/jmlr/RaffelSRLNMZLL20} is used because of its outstanding generation performance. Similar to its original setting, we define the task as a sequence transformation task by adding the prefix ``\textit{translate knowledge into sentence}" at the beginning as $\gP_g$ to guide the generation model.
It is difficult to directly generate text from structured event records with limited training data, 
so we randomly mask the original sentence with the special token \texttt{[M]} to produce the masked sentence $\gC^\prime$, and the mask rate is $\lambda$. $\gC^\prime$ is used as the background in the input of the generation model $\gX_g$.
As shown in Figure~\ref{fig: info2text}, the structured information annotated in the training set is transformed into event description $\gD_g$ and relation description $\gR_g$, respectively. 
They are further used as background knowledge to assist in the structure-to-text generation and the original sentence $\gC$ is regarded as the generation target $\gY_g$.
Given the previously generated tokens $y_{<s}$ and the input $\gX_g$.
It is notable that the entire probability $p(\gY_g \mid \gX_g)$ is calculated as:
\begin{equation}
\begin{aligned}
 p(\gY_g \mid \gX_g) & =\prod_{s=1}^{|\gY_{g}|} p\left(y_{s} \mid y_{<s}, \gX_g\right) \\
 \gX_g & =\left[\gP_g ; \gD_g ; \gR_g ; \gC^\prime \right]
\end{aligned}.
\label{equ: gen}
\end{equation}
In addition, an argument-aware loss $\gL_a$ is added to enforce the model to help the model to pay more attention to the event arguments during the generation process. For all event arguments that have not been generated, we search for text spans in the generated text most similar to the remaining event arguments.
Detailly, 
we aggregate the triggers and arguments which not included in the generated text. These triggers and arguments are transformed into a one-hot embedding set $\sA$ and each element is denoted as $a_m \in A$ denote.
And the probability of selecting the token at each position in the generation model is extracted for matching the optimal-related position.
By setting the window size to the number of words in $\ra_m$, we divide the probability sequence into pieces using the sliding window and obtain all the candidate set $\sK_m$ for each $\ra_m$ in $\sA$.
We first calculate the $L1$ distance between $\ra_m$ and each element in $\sK_m$ as the distance score between them.
Then, all distance scores are mixed together in the back of completely traversing $\sA$.
in the case of avoiding the conflict of matching positions, greedy search is finally utilized to check each element in $\sA$ to the position with the lowest distance score. 
Together with the original language model loss function $\gL_{lm}$, the loss function of the generation model $\gL_g$ is defined as:
\begin{equation}
    \begin{aligned}
    \gL_{lm} &=\sum_{s=1}^{|\gY_g|} y_s log \ p(y_s \mid y_{<s}, \gX_g)\\
    \gL_{a} &= \sum_{t=1}^{\ermT} \sum_{k = k_t}^{k_t^\prime} y_k log \ p(y_k \mid y_{<k}, \gX_{g}) \\
    \gL_{g} & = -\frac{1}{\ermN} \sum_{n=1}^{\ermN} (\beta \gL_{lm} + \gamma \gL_{a} )
    \end{aligned}
\label{equ:gen loss}
\end{equation}
where $\ermN$ is the number of instances, $\ermT$ is the number of elements contained in the current unmatched set, $k_t$ and $k_t^\prime$ denote the start and end position of $t$-th unmatched element in the original sentence, and $y_k$ is the $k$-th corresponding trigger or argument word.

\subsubsection{Policy Model} 
For each input sentence, our policy model is required to determine whether it expresses the target event records. Thus, the policy model makes a removal action if it is irrelevant to the target event records and it is analogous to a binary classifier. 
For each generated sentence $\gG \in \sG_i$, the input of the policy model $\gX_p$ consists of $\gG$ and corresponding event description $\gD_g$. The symbolic representation of input is formulated as $\gX_p = \left[\gD_g ;\texttt{[SEP]} ; \gG \right]$ with the separate marker \texttt{[SEP]}.
We fine-tune the BERT model by feeding the \texttt{[CLS]} vector into the MLP layer. And then a softmax function is utilized to calculate the decision probability for retaining the sample $\gG$.
A binary cross-entropy loss function is introduced for this classifier,
\begin{align}
\gL_p = -\frac{1}{\ermN} \sum_{n=1}^{\ermN}y_n \log p(y_n \mid \gX_p), 
\end{align}
where $y_n$ is the golden action for $n$-th sample, and $\ermN$ is the number of instances.

\subsection{Training Strategy} 
\subsubsection{Pre-training}
The three components, i.e., event extraction model, structure-to-text generation model, and policy model, are pre-trained with different strategies. Since the policy model has no task-specific information at the very beginning, 
the generation model is trained for several epochs at first to establish the training set for the policy model.
We stop training the generation model until more than 70\% of the trigger and arguments could be generated.
The generated sentences containing their corresponding triggers and arguments are considered positive samples for the policy model, while the others are treated as negative samples.
To get a balance between positive and negative samples, 
we randomly select some event descriptions and sentences irrelevant to the event descriptions as negative samples as well.
We early stop training the policy model when the precision reaches $ 80\% \sim 90\%$. 
This can preserve the information entropy of the result predicted by the policy model, and extend the exploration space.
Then we continue to pre-train the generation model and the event extraction model with the original training set for fixed epochs. 
These two pre-trained models are used as our initialized generation model and extraction model in the retraining process, respectively.

\subsubsection{Retraining with Rewards} 

For $i$-th epoch in retraining the agent, the policy model selects actions for each element in generated dataset $\sG_i$. According to the actions, $\sG_i$ is divided into negative samples $\sN_i$ and positive samples set $\sP_i$.
Then we sample a subset from the original training data, and $\sT_o$ is mixed with $\sP_i$ as the reconstructed training set $\sT_i$ and used to retrain the event extraction model.
Except for the improvement of argument F1 score, the growth on trigger F1 is also beneficial for the model.
Therefore, we updated the checkpoint while either the trigger or argument F1 score improved to avoid falling into a local optimum.
Following \cite{qin-etal-2018-robust}, we employ two sets for training the policy model,
\begin{equation}
\begin{aligned}
    \sD_{i-1} & = \sN_{i-1} - (\sN_{i-1} \cap \sN_{i} ) \\
    \sD_i & = \sN_{i} - (\sN_{i-1} \cap \sN_{i} ) \\
\end{aligned}.
\label{equ: policy dataset}
\end{equation}
Since we can't explore all directions to get the maximum reward for a single step, we select a constant number of samples from $\sD_{i-1}$ and $\sD_i$ for training, respectively, named $\sD_{i-1}^\prime$ and $\sD_i^\prime$.
Referring to Equation~(\ref{equ:policy loss}), the retraining loss function of our policy model $\gL_p^\prime$ is defined as:
\begin{equation}
\begin{aligned}
    \gL_p^\prime = & \sum^{\sD_{i}^\prime} y_n \log p(y_n \mid \gX_p) \gR_i + \\
            &\sum^{\sD_{i-1}^\prime}y_n \log p(y_n \mid \gX_p) (-\gR_i).
\label{equ:policy loss}
\end{aligned}
\end{equation}
The probability of being considered an invalid sample is taken as the weight for retraining the corresponding instance in the generation model.
So we use the probability of removing the sample $w_n = 1 - \log p(y_n \mid \gX_p)$ as the sample weight and retrain the generation model with the following retraining loss function $\gL_g^\prime$ referring to Equation~(\ref{equ:gen loss}):
\begin{equation}
     \gL_g^\prime =  -\frac{1}{\ermN} \sum_{n=1}^{\ermN} ( \beta w_n\gL_{lm}^n + \gamma w_n\gL_{a}^n )
\label{equ:retrain gen loss}
\end{equation}
where $\gL_{lm}^n$ and $\gL_{a}^n$ are the language model loss and argument-aware loss for $n$-th sample, respectively.
The detail of the retraining algorithm is shown in Appendix~\ref{sec:rl}.
\section{Experiments}

\subsection{Experimental Settings}
\subsubsection{Datasets and Evaluation Metrics}
Following the previous work~\cite{Zhang:2019:GAIL,wadden-etal-2019-entity,du-cardie-2020-event,lu-etal-2021-text2event,DBLP:journals/corr/abs-2108-12724,DBLP:conf/acl/LiuHSW22}, We preprocess the two widely used English event extraction benchmarks, ACE 2005 (LDC2006T06) and ERE (LDC2015E29, LDC2015E68, and LDC2015E78) into ACE05-E and ERE-EN. ACE 2005 is further preprocessed into ACE05-E$^+$ following~\cite{lin-etal-2020-joint}.
Statistics of the datasets are further shown in Appendix~\ref{sec:ds}.

Following previous work~\cite{Zhang:2019:GAIL,wadden-etal-2019-entity}, we use precision (P), recall (R), and F1 scores to evaluate the performance. More specifically, we report the performance on both trigger classification (\textbf{Trig-C}) and argument classification (\textbf{Arg-C}).
In the task of trigger classification, if the event type and the offset of the trigger are both correctly identified, the sample is denoted as correct.
Similarly, correct argument classification means correctly identifying the event type, the role type, and the offset of the argument.
Following \cite{lu-etal-2021-text2event,DBLP:conf/acl/LiuHSW22}, the offset of extracted triggers is decoded by string matching in the input context one by one. 
For the predicted argument, the nearest matched string is used as the predicted trigger for offset comparison.

\subsubsection{Baselines}
\begin{table}[t]
\small
\centering
\setlength\tabcolsep{1.5pt}
\resizebox{\columnwidth}{!}{
\begin{tabular}{lllllll}
\\
\hline
\multirow{2}{*}{Model} 
& \multicolumn{3}{c}{\textbf{Trg-C}} & \multicolumn{3}{c}{\textbf{Arg-C}}   \\
& \ P       & \ R      & \ F1     & \ P     & \ R     & \ F1 \\
\hline
\textsc{OneIE}                 & \underline{72.1}    & 73.6   & 72.8   & \underline{55.4}  & 54.3  & 54.8 \\
\textsc{Text2Event}             & 71.2    & 72.5   & 71.8   & 54.0  & \underline{54.8}  & 54.4 \\
\textsc{DEGREE-e2e}             & -       & -      & 72.7   & -     & -     & \underline{55.0}\\
\textsc{GTEE-dynpref}             & 67.3    & \textbf{83.0}   & \underline{74.3}   & 49.8  & \textbf{60.7}  & 54.7\\
\textsc{DAEE}  & \textbf{78.8}$^{\pm0.4}$   &\underline{75.1}$^{\pm5.0}$  &\textbf{76.9}$^{\pm0.4}$   &\textbf{58.5}$^{\pm1.5}$   & 54.4$^{\pm0.4}$   &\textbf{56.3}$^{\pm0.2}$ \\
\hline
\end{tabular}
}
\caption{Results on ACE05-E$^+$. We reported the average result of eight runs with different random seeds, our results are like ``$a ^{\pm b}$'', where ``$a$'' and ``$b$'' represents the mean and the variance, respectively. We bold the highest scores and underline the second highest scores.}
\label{tab:e+}
\end{table}

\begin{table}[t]
\small
\centering
\setlength\tabcolsep{1.5pt}
\resizebox{\columnwidth}{!}{
\begin{tabular}{lllllll}
\\
\hline
\multirow{2}{*}{Model} 
& \multicolumn{3}{c}{\textbf{Trg-C}} & \multicolumn{3}{c}{\textbf{Arg-C}}   \\
& \ P       & \ R      & \ F1     & \ P     & \ R     & \ F1 \\
\hline
\textsc{OneIE}   & 58.4     & 59.9    & 59.1      & \underline{51.8}    & \underline{ 49.2}   & 50.5  \\
\textsc{Text2Event}           & 59.2       & 59.6      & 59.4      & 49.4    & 47.2   & 48.3\\
\textsc{DEGREE-e2e}             & -          & -         & 57.1      & -       & -      & 49.6   \\
\textsc{GTEE-dynpref}             & \underline{61.9}       & \textbf{72.8 }     & \textbf{66.9}      & 51.9    & \textbf{58.8}   & \textbf{55.1}\\
\textsc{DAEE}  &\textbf{68.7}$^{\pm0.8}$    &\underline{61.6}$^{\pm0.5}$  &\underline{65.0}$^{\pm0.4}$  &\textbf{57.7}$^{\pm0.8}$  &46.7$^{\pm0.4}$   &\underline{51.6}$^{\pm0.3}$     \\
\hline
\end{tabular}
}
\caption{Results on ERE-EN.}
\label{tab:ere}
\end{table}

\begin{table}[t]
\small
\centering
\setlength\tabcolsep{1.5pt}
\resizebox{\columnwidth}{!}{
\begin{tabular}{lllllll}
\\
\hline
\multirow{2}{*}{Model}                  & \multicolumn{3}{c}{\textbf{Trg-C}} & \multicolumn{3}{c}{\textbf{Arg-C}} \\
                                        & \ P       &\ R      &\ F1     &\ P       &\ R      &\ F1     \\ \hline
\textsc{DyGIE++}                                 & -       & -      & 69.7   & -       & -      & 48.8   \\
\textsc{GAIL}                                    & \underline{74.8}    & 69.4   & 72.0   & \textbf{61.6}    & 45.7   & 52.4   \\
\textsc{OneIE}                                   & -       & -      & \underline{74.7}   & -       & -      & \textbf{56.8}   \\
\textsc{BERT\_QA}                                & 71.1    & 73.7   & 72.3   & \underline{56.8}    & 50.2   & 53.3   \\
\textsc{MQAEE}                                   & -       & -      & 71.7   & -       & -      & 53.4   \\ \hline
\textsc{TANL}                                    & -       & -      & 68.5   & -       & -      & 48.5   \\
\textsc{BART-Gen}                               & 69.5    & 72.8   & 71.1   & 56.0    & 51.6   & 53.7   \\
\textsc{Text2Event}                              & 67.5    & 71.2   & 69.2   & 46.7    & 53.4   & 49.8   \\
\textsc{DEGREE-e2e}                              & -       & -      & 70.9   & -       & -      & 54.4   \\
\textsc{GTEE-dynpref}                            & 63.7    & \textbf{84.4}   & 72.6   & 49.0    & \textbf{64.8}   & 55.8   \\

\textsc{DAEE}  &\textbf{75.1}$^{\pm1.7}$   &\underline{76.6}$^{\pm4.1}$   &\textbf{75.8}$^{\pm0.6}$    &55.9$^{\pm3.6}$     &\underline{57.2}$^{\pm1.8}$    &\underline{56.5}$^{\pm0.3}$ \\
\hline
\end{tabular}
}
\caption{Results on ACE05-E. The first group is the classification-based methods and the second group is the generation-based methods.}
\label{tab:ace05-e}
\end{table}

We illustrate the event extraction results between our proposed \textsc{DAEE} and the baselines conducted in two categories, i.e., classification-based models and generation-based models.

The first category is \textbf{classification-based models},
\textsc{DyGIE++} \cite{wadden-etal-2019-entity}: a joint model with contextualized span representations.
\textsc{GAIL} \cite{Zhang:2019:GAIL}: an RL model jointly extracting entity and event.
\textsc{OneIE} \cite{lin-etal-2020-joint}: a joint neural model for information extraction task with several global features and beam search.
\textsc{BERT\_QA} \cite{du-cardie-2020-event}: a method using separated question-answering pairs for event extraction.
\textsc{MQAEE} \cite{li-etal-2020-event}: a question answering system with multi-turn asking.

The other category is \textbf{generation-based methods}, and our proposed \textsc{DAEE} belongs to this one.
\textsc{TANL} \cite{DBLP:conf/iclr/PaoliniAKMAASXS21}: a method that use translation tasks modeling event extraction in a trigger-argument pipeline.
\textsc{BART-Gen} \cite{li-etal-2021-document}: a document-level event extraction method through conditional generation.
\textsc{Text2Event} \cite{lu-etal-2021-text2event}: 
a method directly generates structure from the text.
\textsc{DEGREE-e2e} \cite{hsu-etal-2022-degree}: a method using discrete prompts and end-to-end conditional generation to extract event.
\textsc{GTEE-dynpref} \cite{DBLP:conf/acl/LiuHSW22}: a generative template-based event extraction method using dynamic prefix-tuning. 

\subsection{Results and Analysis}
\subsubsection{Main Results}

The performance comparison on dataset ACE05-E$^+$ is shown in Table~\ref{tab:e+}. It can be observed that \textsc{DAEE} achieves the SOTA F1 score on ACE05-E$^+$ and obtain 1.1\% and 0.7\% gain of F1 scores for\textbf{Trg-C} and \textbf{Arg-C}, respectively.
The improvement indicates that \textsc{DAEE} is able to guide the generation model to generate the text containing events and select suitable samples to improve the effectiveness of the event extraction model.

Table~\ref{tab:ere} presents the performance of baselines and \textsc{DAEE} on ERE-EN. The performance of \textsc{DAEE} decreases compared with \textsc{GTEE-dynpref}, but the performance is still higher than other methods, which may be affected that ERE-EN contains more pronoun arguments. The pronoun roles would offer less information for the generation model thus reducing the role of structured text in guiding the generation model.

Comparing the results on ACE05-E as Table~\ref{tab:ace05-e} shows, we gain an improvement of $1.1\%$ on \textbf{Trg-C} and a competitive F1 score on \textbf{Arg-C} with the SOTA classification-based method \textsc{OneIE}, outperforming the others.
This observation supports that structured information used in the knowledge-based generation model makes up for the information gap used by multi-task extraction.

\subsubsection{Ablation Study}
\begin{table}[t]
\small
\centering
\begin{tabular}{lcccccc}
\hline
\multicolumn{1}{l}{\multirow{2}{*}{Model}} & \multicolumn{3}{c}{\textbf{Trg-C}} & \multicolumn{3}{c}{\textbf{Arg-C}}      \\
\multicolumn{1}{c}{}                       & P       & R      & F1     & P     & R     & F1             \\ \hline
\textsc{DAEE}   &\underline{78.8}   &75.1  &\underline{76.9}   &58.5    & 54.4   &\textbf{56.3} \\
\quad w/o \textsc{AL}  & 78.0  &\textbf{75.5} &76.7  & 56.2  &\textbf{55.6} &\underline{55.9}  \\
\quad w/o \textsc{RG}   &78.8 &\underline{75.5} & \textbf{77.1}  &\underline{56.2}  & \underline{54.9} & 55.5 \\
\quad w/o \textsc{RL}  &\textbf{79.0} &71.9 &75.3 & \textbf{56.3} &54.3 & 55.3 \\
\hline
\end{tabular}
\caption{Ablation Study on ACE05-E$^+$ for event extraction. \textsc{AL} denotes the argument-aware loss $\gL_a$,  \textsc{RG} denotes the process of retraining the generation model, and \textsc{RL} denotes the reinforcement learning strategy.}
\label{tab:ablation}
\end{table}
We further conducted an ablation study by removing each module at a time. 
The experimental results on ACE05-E$^+$ are presented in Table \ref{tab:ablation}.
We can see that the F1 score of \textbf{Arg-C} decreases by $0.4\%$ and $0.8\%$ when removing the argument-aware loss $\gL_a$ and stopping retraining the generation model, respectively.
The results indicate that the deployment of argument-aware loss and retraining strategy is conducive to the generation module in our framework.
Then, we remove the RL strategy, which means that the generated samples are directly mixed with the original training samples for training the event extraction model from scratch.
The F1 score of \textbf{Trg-C} and \textbf{Arg-C} decreases by $1.6\%$ and $1.0\%$, respectively.
This demonstrates that the RL strategy could ensure that the generated data is more suitable for downstream event extraction tasks and guide the improvement on both \textbf{Trg-C} and \textbf{Arg-C}.

\subsubsection{Iterative Generation Discussion} 
\begin{figure}[t]
\centering
\includegraphics[width=0.95\columnwidth]{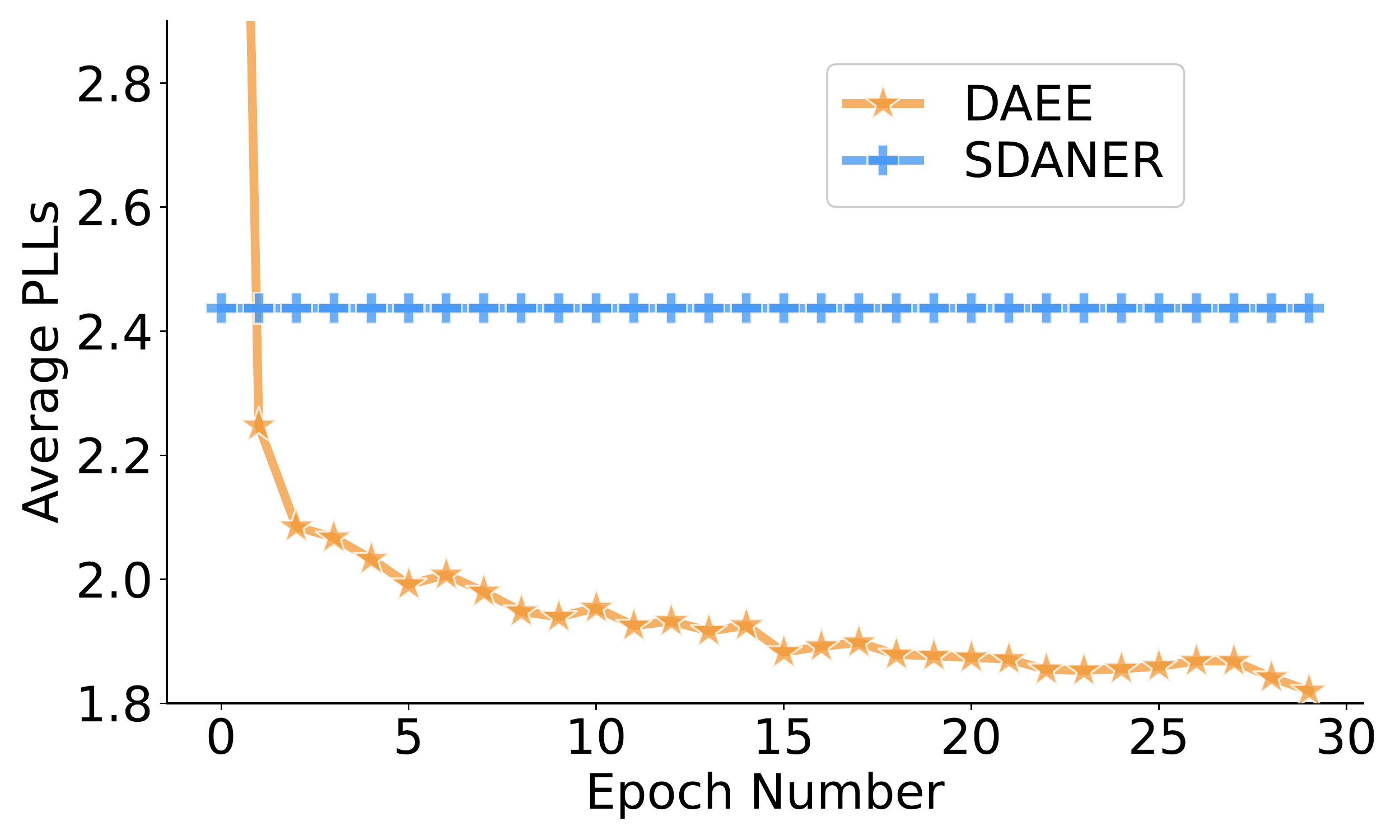} 
\caption{Results for the PLLs of \textsc{DAEE} and \textsc{SDANER} on ACE05-E$^+$.}
\label{fig: gen-it}
\end{figure}

To illustrate our framework is able to enhance the quality of generated sentences, we calculate the masked language model score \textit{pseudo-log-likelihood scores}~(PLLs)\footnote{
BERT is fine-tuned through mask language model loss using the training set for calculating PLLs.} following \cite{salazar-etal-2020-masked} for each training epoch.
The token $\vw_s$ in the sentence is masked and predicted using all past and future tokens $\mW_{\backslash s}:= (\vw_1, \dotsc, \vw_{s-1}, \vw_{s+1}, \dotsc, \vw_{\abs{\mW}})$, and the PLLs for each sentence is calculated as
\begin{align*}
    \text{PLLs}(\mW) :=  \frac{1}{\abs{\mW}} \sum_{t=1}^{\abs{\mW}} \log P_{\text{MLM}}(\vw_s \mid \mW_{\backslash s}; \Theta).
\end{align*}
The results for each epoch are the average of sentence scores over the entire training set as shown in Figure~\ref{fig: gen-it}.
PLLs is declining with the iterative process, which demonstrates that DAEE enhances the fluency of generated data and improves the effect of event extraction under the guidance of RL agent.
Furthermore, we compare \textsc{DAEE} with a rule-based sequence labeling data augment method \textsc{SDANER}~\cite{DBLP:conf/coling/DaiA20}. \textsc{SDANER} contains four rule-based augmentation methods. Synonym replacement is selected according to its lowest average PLLs. \textsc{DAEE} generates sentences with lower PLLs compared with the rule-based method. The results demonstrate that \textsc{DAEE} generates more fluency and grammatically correct data.

\begin{table*}[t]
\small
\centering
\setlength{\belowcaptionskip}{-0.1cm}
\begin{tabular}{ll}
\\ \hline
Event type & Transaction:Transfer-Ownership \\ 
Original sentence & yes, \textcolor[rgb]{0.047, 0.412, 0.004}{\textbf{we}} got uh \textcolor[rgb]{0.792,0,0.012}{\textbf{purchased}} by our strategic \textcolor[rgb]{0.047, 0.412, 0.004}{\textbf{partner}}, so um\\
\textsc{Generation Model} (w/o $\gL_a$) &  yeah , \textcolor[rgb]{0.047, 0.412, 0.004}{\textbf{we}} bought from our \textcolor[rgb]{0.047, 0.412, 0.004}{\textbf{partner}}, um, um \\ 
\textsc{Generation Model} &  well , \textcolor[rgb]{0.047, 0.412, 0.004}{\textbf{we}} \textcolor[rgb]{0.792,0,0.012}{\textbf{purchased}} our \textcolor[rgb]{0.047, 0.412, 0.004}{\textbf{partner}} purchased, um \\
\textsc{DAEE} &  yeah, \textcolor[rgb]{0.047, 0.412, 0.004}{\textbf{we}} got uh \textcolor[rgb]{0.792,0,0.012}{\textbf{purchased}} by our \textcolor[rgb]{0.047, 0.412, 0.004}{\textbf{partner}},\\ 
\hline
Event type & Life:Die  \&  Conflict:Attack \\ 
Original sentence & the iraqi government reports 1252 civilians have been \textcolor[rgb]{0,0, 0.843}{\textbf{killed}} in \textcolor[rgb]{0.792,0,0.012}{\textbf{the war}}.\\
\textsc{Generation Model} (w/o $\gL_a$)&  the iraqi government says more than 200 \textcolor[rgb]{0.047, 0.412, 0.004}{\textbf{civilians}} have been \textcolor[rgb]{0,0, 0.843}{\textbf{killed}} in this war . \\ 
\textsc{Generation Model} &  the iraqi government \textcolor[rgb]{0,0, 0.843}{\textbf{killed}} \textcolor[rgb]{0.047, 0.412, 0.004}{\textbf{civilians}} in \textcolor[rgb]{0.792,0,0.012}{\textbf{the war}} . \\
\textsc{DAEE}&  the iraqi government says more than 200 \textcolor[rgb]{0.047, 0.412, 0.004}{\textbf{civilians}} have been \textcolor[rgb]{0,0, 0.843}{\textbf{killed}} \textcolor[rgb]{0.792,0,0.012}{\textbf{the war}} .\\ 
\hline
\end{tabular}
\caption{Efficient generated synthetic data from our proposed methods and simple generated Sentence. Text chunks in Blue and Red are the event triggers for different event type, text chunks in Green are the event arguments.}
\label{tab:case study}
\end{table*}

\begin{figure}[t]
\centering
\includegraphics[width=0.92\columnwidth]{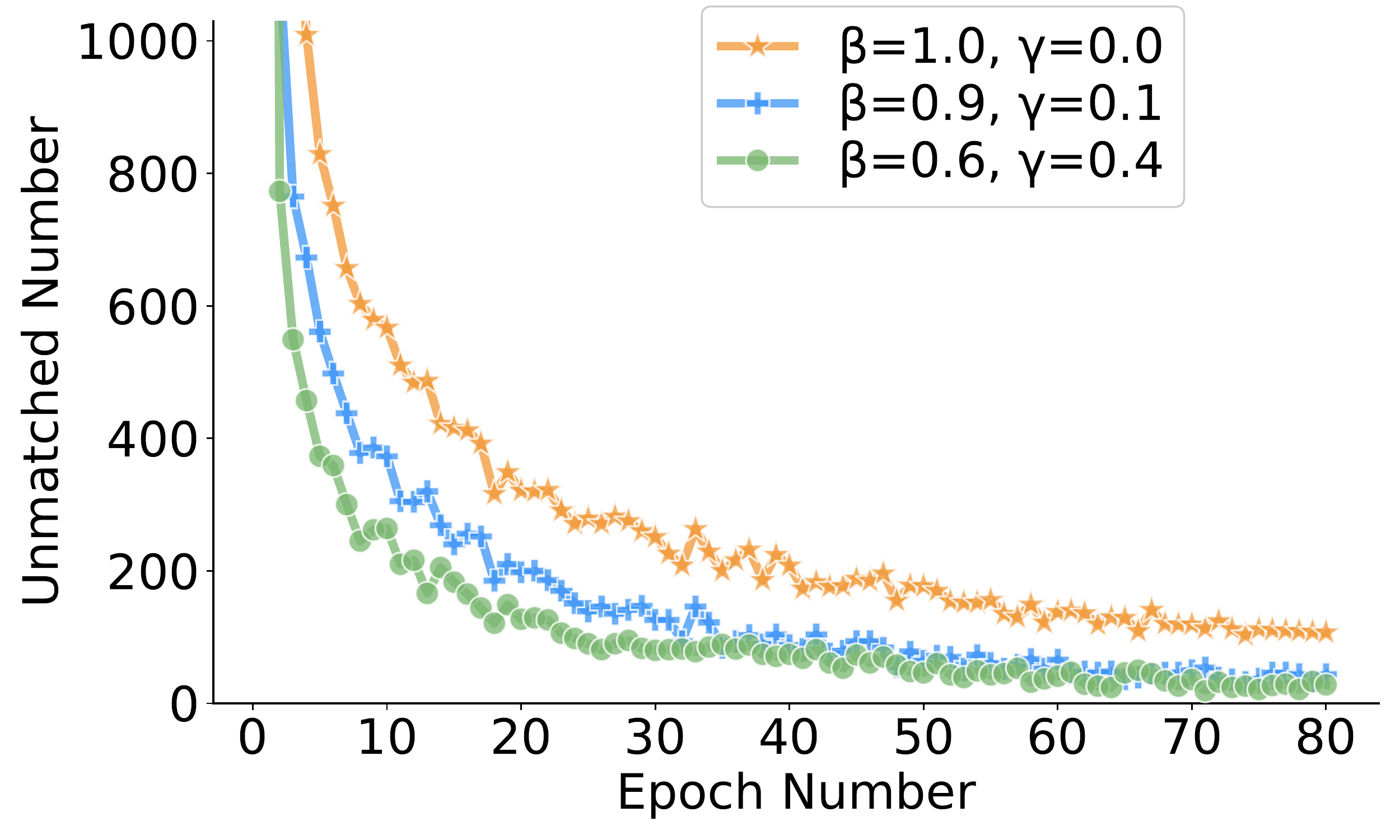} 
\caption{Unmatched arguments numbers of different training epochs.}
\label{fig: unmatch}
\end{figure}
\subsubsection{Argument Loss Analysis}
To verify the effectiveness of argument-aware loss $\gL_a$ in reducing mismatches triggers and arguments, we alter the hyperparameter $\gamma$ and explore the change of the unmatched number of arguments during the training process.
Three generation models are trained according to the loss function mentioned in Equation~(\ref{equ:gen loss}), and results shown in Figure \ref{fig: unmatch} are observed by the change in the ratio of $\beta$ and $\gamma$.
Compared with setting $\gamma$ to $0$, the number of unmatched arguments drops rapidly under the condition of adding the $\gL_a$ by increasing the $\gamma$. Meanwhile, the number of unmatched arguments converges around $30$ after adding $\gL_a$, while the number converges to around $120$ without $\gL_a$. 

\subsubsection{Diversity Analysis} Intuitively, diverse sentence description in the training set is able to enhance the model performance. We thus verify the diversity of the generated text. The degree of diversity is reported by calculating the number of distinct bigrams and trigrams in the generated text which has not appeared in the original text and the results are shown in Table \ref{tab:diversity}.
In the following, we use \textsc{Generation Model} to represent the directly trained structure-to-text generation model.
Referring to the indicators proposed in \cite{DBLP:conf/naacl/LiGBGD16}, 
The diversity, the argument-aware loss $\gL_a$ helps the \textsc{Generation Model} to produce more diverse synthetic data, which is because the argument-aware loss makes the model focus more on retaining the triggers and arguments rather than generating more similar content to the original text.
The diversity is affected by the RL strategy due to the concentration on the effect of event extraction.
Horizontally compared to Table~\ref{tab:ablation}, 
the experimental results demonstrate that diversified text can enable the model to obtain more information based on similar event records.

\subsubsection{Synthetic Data Case Study} Table \ref{tab:case study} shows representative examples generated by our proposed \textsc{DAEE} and other methods and we can see the following comparative phenomena.
In the case of comparing whether to add the argument-aware loss, the \textsc{Generation Model} generates all the triggers and arguments in three examples, which demonstrate the generation model without $\gL_a$ shuffles the text leaking problem.
There is a misalignment in the first example for the text generated through \textsc{Generation Model}. The original sentence contains two roles, i.e., \textsc{Artifact} and \textsc{Buyer}, and their arguments are \textit{we} and \textit{partner}, but the two arguments have been swapped in the synthetic text.
In the second example, the \textit{government} should play the role of \textsc{Agent} in \textsc{Life:Die} event according to the output of \textsc{Generation Model}, which is not appeared in the golden event record and resulting in redundancy.
Neither of the above errors occurs in \textsc{DAEE} shown in the table, which proves the RL strategy could also be guidance for improving the effectiveness of generative models.
\begin{table}[t]
\small
\centering
\setlength{\belowcaptionskip}{-0.3cm}
\begin{tabular}{lccc}
\hline
Model  & bigrams & trigrams\\
\hline
\textsc{Generation Model}  &0.160 & 0.398\\
\textsc{Generation Model} (w/o $\gL_a$)  &0.125 & 0.323\\
\textsc{DAEE}    & 0.143 &  0.365\\
\hline
\end{tabular}
\caption{Results of diversity analysis on ACE05-E$^+$.}
\label{tab:diversity}
\end{table}
\section{Conclusion}
In this paper, we studied \textsc{DAEE}, the denoised structure-to-text augmentation framework for event extraction. 
The structure-to-text generation model with argument-aware loss is guided by the reinforcement learning agent to learn the task-specific information.
Meanwhile, the reinforcement learning agent selects effective samples from generated training data that are used to reinforce the event extraction performance.
Experimental results show that our model achieves competitive results with the SOTA on ACE 2005, which is also a proven and effective generative data augmentation method for complex structure extraction.
\section{Limitation}
This paper proposes a denoised structure-to-text augmentation framework for event extraction (\textsc{DAEE}), which generates and selects additional training data iteratively through RL framework. However, we still gain the following limitations.
\begin{itemize}
    \item The framework uses reinforcement learning to select effective samples, which is a process of iterative training and predicting the generation model, policy model, and event extraction models.
    The iterative training framework is complicated and time-consuming compared to the standalone event extraction model.
    \item  Even the Argument Loss decreases the number of unmatched arguments in a generated sentence, the generation model generates more fluent sentences while at the expense of the ability to ensure that all the event arguments are included completely.
\end{itemize}


\section{Acknowledgement}
This work was supported by the Joint Funds of the National Natural Science Foundation of China (Grant No. U19B2020).
We would like to thank the anonymous reviewers for their thoughtful and constructive comments.

\bibliography{custom}
\bibliographystyle{acl_natbib}

\newpage
\appendix
\section{Details of Methods}
The detail of the retraining algorithm is shown in Algorithm~\ref{alg: rl}.
\label{sec:rl}
\begin{algorithm*}[tp]
\caption{The process of retraining the reinforcement learning framework.}
\label{alg: rl}
\textbf{Parameter}:The original event extraction training set $\sT_o$, parameters of policy model $\theta_p$, event extraction model $\theta_e$, generation model $\theta_g$, generated sentence set, $n$-th generated sentence $\gG_n$, positive samples set $\sP_i$, negative samples set $\sN_i$
\begin{algorithmic}[1] 
\STATE Initialize trigger $F1$ score $F_{max}^t$ and role $F1$ score $F_{max}^a$ through $\theta_e$
\FOR{epoch $i$ in $1 \to \ermK $}
\FOR{$\gG_n$ in $\sG_{i-1}$}
\STATE Calculate $\left[\gD_g ;[S E P] ; \gG_n \right]  \to \gX_p$ 
\STATE Sample action according $p(y_n \mid \gX_p,\theta_p)$ 
\IF{action == 1}
\STATE Add $\gG_{n} \to \sP_i$
\ELSE
\STATE Add $\gG_{n} \to \sN_i$
\ENDIF
\ENDFOR
\STATE Calculate $\sD_i^\prime$ and $\sD_i^\prime$ according Equation \ref{equ: policy dataset}
\STATE Sample $\sT_{sub}$ from $\sT_o$ and concatnate $\{\sT_{sub}, \sP_i\} \to \sT_i$
\STATE Retrain event extraction model through $\sT_i$
\STATE Calculate \textbf{Trg-C} score $F_i^t$ and \textbf{Arg-C} score $F_i^a$, training set for \textsc{Generation Model} $\sY_i$
\STATE Calculated Reward $\alpha(F_i^a - F_{i-1}^a) \to \gR_i$
\IF{$F_i^a > F_{max}^a$ or $F_i^t > F_{max}^a$}
\STATE Change $F_i^a  \to F_{max}^t$, $F_i^t  \to F_{max}^t$, and update $\theta_p$
\ENDIF  
\STATE Retrain policy through $\sD_i$ and $\sD_{i-1}$ according Equation \ref{equ:policy loss}
\STATE Update training weight $1-\log p(\gY_p \mid \gX_p) \to w_n$ for each sample in $\sY_g$, 
\STATE Retrain the generation model through weighted $\sY_g$ according Equation \ref{equ:gen loss}
\STATE Update $\theta_g$ and generate $\sG_i$
\ENDFOR
\end{algorithmic}
\end{algorithm*}

\section{Details of Experiments}
\subsection{Data Statistics}
\label{sec:ds}
In this paper, we use the three datasets to verify our proposed method, the statistics of the datasets are shown in Table~\ref{tab:ds}.
\begin{table}[t]
\small
\centering
\begin{tabular}{ccccc}
\hline
Dataset  & Split & \#Sents & \#Events & \#Roles \\ \hline
         & Train & 17,172  & 4,202    & 4,859   \\
ACE05-E  & Dev   & 923     & 450      & 605     \\
         & Test  & 832     & 403      & 576     \\
         & Train & 19,216  & 4,419    & 6,607   \\
ACE05-E$^+$ & Dev   & 901     & 468      & 759     \\
         & Test  & 676     & 424      & 689     \\
         & Train & 14,736  & 6,208    & 8,924   \\
ERE-EN   & Dev   & 1,209   & 525      & 730     \\
         & Test  & 1,163   & 551      & 822     \\ \hline
\end{tabular}
\caption{Dataset statistics.}
\label{tab:ds}
\end{table}

\subsection{Implementation Details}
All experiments were conducted with NVIDIA A100 Tensor Core GPU 40GB.
For the pre-trained language model, we reuse the three English models released by Huggingface\footnote{\url{https://huggingface.co/t5-base},\\ \url{https://huggingface.co/bert-base-uncased}, \\ \url{https://huggingface.co/facebook/bart-large}}. 
Specifically, $\gamma$ and $\beta$ are set to $0.1$ and $0.9$ in Equation~(\ref{equ: gen}), respectively, the RL training epoch is set to 80, the reward scale $\alpha$ is set to 10, the sample ratio from original event extraction training set is set to 0.5, the negative sample ratio for \textsc{GTEE-base} in training is set to 12\% for event extraction, and the other hyperparameters used are shown in Table \ref{tab:hyp}.

\begin{table}[t]
\centering
\small
\begin{tabular}{lccc}
\hline
Name & \textsc{EE} & \textsc{Policy} & \textsc{GEN} \\
\hline
learning rate (pretrain) & 1e-5 & 1e-5 & 3e-5 \\
learning rate (retrain) & 1e-6 & 1e-6 & 3e-5 \\
train batch size & 32*2 & 32 & 32 \\
epochs (pretrain) & 15 & - & 20 \\
epochs (retrain) & 2 & 1 & 1 \\
weight decay (pretrain) & 1e-5 & 1e-5 & 1e-5 \\
gradient clip & 5.0 & 5.0 & 5.0 \\
warm-up ratio (pretrain) & 10\% & - & - \\
optimizer & AdamW & Adam& Adam \\        
\hline
\end{tabular}
\caption{Hyperparameter setting for our models, 
\textsc{EE} denotes the event extraction model, \textsc{Policy} denotes the policy model, \textsc{GEN} denotes the generation model.}
\label{tab:hyp}
\end{table}%

\subsection{Generation Reliability Discussion}
To verify the verifies the convince of the generated data, we train \textsc{GTEE-base} through the samples with event record, which is because that only the samples with event record are used for data augmentation. The results are shown in Table~\ref{tab:ablation-base}. 
The $F1$ score trained on \textsc{DD} increases by $1.1\%$ and $2.5\%$ compared with the results trained on \textsc{OD} and \textsc{GD}, respectively.
The data generated by \textsc{DAEE} achieves a closer effect to original data, which thus could be utilized for training the competitive event extraction models.
\begin{table}[t]
\small
\centering
\begin{tabular}{lcccccc}
\hline
\multicolumn{1}{c}{\multirow{2}{*}{Model}} & \multicolumn{3}{c}{\textbf{Trg-C}} & \multicolumn{3}{c}{\textbf{Arg-C}}      \\
\multicolumn{1}{c}{}                       & P       & R      & F1     & P     & R     & F1             \\ \hline
\textsc{DD}  & 69.3& 79.7 &  74.1&47.6 &56.5 &51.7 \\
\textsc{GD} & 68.5 &  81.4 &  74.4& 42.3 &58.6 &49.2 \\
\textsc{OD} & 66.3 & 80.7&  72.8&43.1 & 61.2 & 50.6   \\
\hline

\end{tabular}
\caption{The experimental results on ACE05-E$^+$,\textsc{DD} denotes using the generated data though \textsc{DAEE}, while \textsc{GD} denotes the data from \textsc{Generation Model} without RL, \textsc{DD} denotes the data from the original training set.}
\label{tab:ablation-base}
\end{table}

\end{document}